\pdfoutput=1
\documentclass[11pt,a4paper]{article}
\usepackage[utf8]{inputenc}
\usepackage[T1]{fontenc}
\usepackage{lmodern}
\usepackage{microtype}
\usepackage[margin=2.5cm]{geometry}
\usepackage{booktabs}
\usepackage{graphicx}
\usepackage{url}
\usepackage{amsmath}
\usepackage{xcolor}
\usepackage[hidelinks]{hyperref}
\usepackage[square,numbers,sort&compress]{natbib}

\newcommand{\ltag}[1]{\texttt{<2#1>}}

\title{Teaching a Multilingual Translation Model Regional Varieties:\\
Castilian and Latin American Spanish, Asturian, Aragonese and Aranese\\
with MADLAD-400-3B-MT}

\author{Nerea de la Torre Veguillas$^{1}$ \qquad Pol Gerdt$^{1}$ \qquad Raul Quijada$^{1}$\\
  Ivory Dipasupil$^{1}$ \qquad Lucía Rado$^{2}$ \qquad Francesc Tarrés$^{1}$\\[2ex]
  $^{1}$Ugiat Technologies, \texttt{ugiat@ugiat.com}\\
  $^{2}$Corporación de Radio y Televisión Española (RTVE)}
\date{}

\begin{document}
\maketitle

\begin{abstract}
Massively multilingual translation models usually have one tag per language and no control over regional varieties,
and they cover a region's smaller languages poorly or not at all. We report three fine-tunes of
MADLAD-400-3B-MT, each adding target tags with low-rank adaptation (LoRA), back-translation of monolingual text and
replay of the base model's own translations: \ltag{es\_ES} for Spanish as written in Spain, \ltag{es\_419} for Latin
American Spanish, and one model for three Romance languages of the Iberian Peninsula, Asturian (\ltag{ast}),
Aragonese (\ltag{an}) and Aranese (\ltag{oc\_aranes}). For the Spanish varieties we annotate the 250.9M documents of
MADLAD-400's Spanish split with a domain-supervised Spain vs.\ Latin America classifier and select variety text with
a rule that admitted no Latin American document in a 340-document reference set. The Castilian model chooses Spain's
everyday vocabulary in 97\% of a targeted test (base model 62\%) and \emph{vosotros} for informal plural address in
70\% (10\%), while improving chrF++ on human references. A first finding is \emph{tag collapse}: with a new tag
initialised from \ltag{es}, the parent tag drifts to the new variety (84--87\% identical outputs). In
the Iberian model, contrastive pairs and a separate learning rate for the tag embeddings keep each tag in its own
language. Trained only on publicly available data, mostly the TAN-IBE Wikipedia corpora, the Iberian model matches
or beats the TAN-IBE Marian systems in four of six Spanish directions on FLORES+, is 1.9 BLEU behind in
Spanish$\to$Aranese, and is second only to SalamandraTA-7B in most directions. The exception, Spanish$\to$Aragonese, reflects a difference of written
variety between Aragonese Wikipedia and the FLORES+ reference, not model quality. All three fine-tunes lose
0.6--0.7 chrF++ on average in thirteen other FLORES-200 directions. We release the models, data and variety classifier
under open licences.
\end{abstract}

\section{Introduction}

Multilingual neural machine translation (MT) systems select the output language with a tag on the source
\citep{johnson2017google}. MADLAD-400-3B-MT \citep{kudugunta2023madlad400}, a 3B-parameter T5 model trained on a web
corpus covering more than 400 languages, follows this design with tags such as \ltag{es} or \ltag{fr}. A single tag
per language hides regional varieties. MADLAD's \ltag{es} writes a Latin American-leaning Spanish that mixes varieties
(\emph{Dejé mi teléfono celular en el coche} uses Latin American \emph{celular} and Peninsular \emph{coche}), which
is wrong for content produced in Spain, and it occasionally answers in \emph{vosotros} forms, which are wrong for
Latin America. For the minority languages of Spain the gap is larger: MADLAD has no Asturian tag, its \ltag{an}
(Aragonese) often answers in Spanish, and its \ltag{oc} produces standard (Languedocien) Occitan, not Aranese, the
variety of Occitan that is co-official in Catalonia.

This report collects our work on three fine-tunes of MADLAD-400-3B-MT for Ugiat Technologies, a company producing
transcription, translation and dubbing for broadcasters in Spain:
\begin{itemize}
\item a \textbf{Castilian Spanish} model with a new tag \ltag{es\_ES} (Section~\ref{sec:spanish}), together with a
  domain-supervised Spain vs.\ Latin America document classifier, a document-level annotation of the whole Spanish
  split of MADLAD-400 (250.9M documents) and a corpus of 18.7M Castilian sentences;
\item its \textbf{Latin American Spanish} counterpart with \ltag{es\_419} (Section~\ref{sec:spanish}), so that neither
  variety is presented as the default;
\item one model for \textbf{Asturian, Aragonese and Aranese} (Section~\ref{sec:iberian}), in honour of the TAN-IBE
  project \citep{tanibe}, whose openly published corpora made the Asturian and Aragonese parts possible.
\end{itemize}
The three share one recipe (Section~\ref{sec:method}): a new target tag stored as a SentencePiece piece and
initialised from the closest existing tag, LoRA adapters \citep{hu2022lora}, back-translated human text in the
target variety \citep{sennrich2016backtranslation}, and replay of the base model's own translations to limit
forgetting. Along the way we found two practical issues that we think are of general interest: \emph{tag collapse},
where the parent tag is pulled along with the new one (Section~\ref{sec:collapse}), and a \emph{written-variety
mismatch} between the Aragonese FLORES+ reference and Aragonese Wikipedia, which changes the ranking of systems
(Section~\ref{sec:variety-mismatch}).

\section{Related work}

Translating into language varieties has been addressed with variety tags on top of a multilingual model
\citep{lakew2018varieties}, by adapting to low-resource varieties with little parallel data
\citep{kumar2021varieties}, and with region-aware benchmarks such as FRMT for Brazilian vs.\ European Portuguese and
Mainland vs.\ Taiwan Mandarin \citep{riley2023frmt}. Back-translation \citep{sennrich2016backtranslation}, optionally
with a tag marking synthetic sources \citep{caswell2019tagged}, is the standard way to use monolingual target text.
Fine-tuning a multilingual model on a few directions degrades the others (catastrophic forgetting,
\citealp{mccloskey1989catastrophic}); replaying data of the original task is the simplest remedy.

For the languages of Spain, the WMT 2024 shared task on translation into low-resource languages of Spain
\citep{sanchezmartinez2024wmt} covered Aragonese, Aranese and Asturian from Spanish, and introduced the PILAR corpora
\citep{pilar}. Rule-based Apertium systems \citep{forcada2011apertium} exist for these pairs and have been used to
create synthetic training data. The TAN-IBE project \citep{tanibe} published Marian models between Spanish and
Asturian, Aragonese and Aranese, together with Wikipedia-based corpora
\citep{tanibe_ast,tanibe_arg,tanibe_arn}. SalamandraTA \citep{garcia-gilabert-etal-2025-salamandra,salamandrata7b,salamandrata2b} and IbRo-nllb
\citep{ibro}, a fine-tune of NLLB \citep{nllb2022}, also cover these languages. FLORES+ \citep{floresplus}, the
continuation of FLORES-101/200 \citep{goyal2022flores,nllb2022}, provides test sets for all of them.

\section{Method}
\label{sec:method}

\paragraph{Base model and tags.} We fine-tune \texttt{google/madlad400-3b-mt}. MADLAD's language tags are
user-defined SentencePiece pieces, encoded as a word-initial \texttt{\_} (U+2581) followed by the tag piece. A new tag must be
added in the same way: added as a transformers ``added token'' instead, it loses that marker, and the untrained
model produces garbage. We add each new tag to the SentencePiece model (ids 256000 and up). Its embedding is a copy
of a related tag's embedding plus a trainable offset (\ltag{es\_ES} and \ltag{es\_419} from \ltag{es}, \ltag{oc\_aranes}
from \ltag{oc}, \ltag{ast} from \ltag{es}), so before training the new tag translates exactly like its parent.
Aragonese reuses MADLAD's own \ltag{an}.

\paragraph{Adaptation.} LoRA \citep{hu2022lora} with rank 16, $\alpha=32$ and dropout 0.05 on the attention ($q,k,v,o$)
and feed-forward ($w_{i0}, w_{i1}, w_o$) projections of encoder and decoder: 47.2M trainable parameters (1.6\%), plus
the tag offsets. One epoch, 128 pairs per step (16 $\times$ 8 gradient accumulation), learning rate $2\cdot10^{-4}$
with cosine decay and 3\% warm-up, AdamW, bf16, gradient checkpointing, sequences of at most 160 tokens. The adapters
are merged into the weights for release, and the merged model is converted to CTranslate2 int8 \citep{ctranslate2}.
Every run used one NVIDIA RTX 4090 (24\,GB) and took 17--22 hours.

\paragraph{Training targets are human text.} For every new tag, the target side is text written by people in that
variety; machine translation appears only on the source side (back-translation). We use the base model in
CTranslate2 (beam 2) to translate target-variety sentences into the source languages, and drop empty, copied,
looping, wrong-language (fastText \texttt{lid.176}, \citealp{joulin2016fasttextzip}) and badly length-matched outputs.
No output of a large language model is used as training data.

\paragraph{Replay.} 25\% of the training mix are \emph{replay} pairs whose targets are the base model's own
translations into 20 other languages (en, fr, de, it, pt, ca, gl, eu, nl, pl, ru, uk, ro, el, tr, ar, zh, ja, ko, hi)
and into plain \ltag{es}, so that the other directions keep their behaviour.

\paragraph{Contrastive pairs and tag learning rate (Iberian model).} To keep a new tag and its parent apart
(Section~\ref{sec:collapse}), the Iberian model adds \emph{contrastive pairs}: the same source sentence paired once
with the new tag and the human target (\ltag{ast}/\ltag{an} $\to$ Asturian/Aragonese, \ltag{oc\_aranes} $\to$ Aranese)
and once with the parent tag and the base model's translation (\ltag{es} $\to$ Spanish, \ltag{oc} $\to$ standard
Occitan). The tag offsets get their own learning rate, $2\cdot10^{-3}$, ten times the LoRA rate.

\paragraph{Evaluation.} BLEU \citep{papineni2002bleu}, chrF \citep{popovic2015chrf} and chrF++ \citep{popovic2017chrfpp}
with sacrebleu defaults \citep{post2018sacrebleu} (spBLEU for FLORES-200); beam 4 for our models. Forgetting is
measured on thirteen FLORES-200 devtest directions (en$\to$fr, de, pt, it, ca, ru, zh, ja, ar, hi; es, fr,
de$\to$en). We check every training corpus against FLORES dev and devtest for exact matches and shared 13-word
sequences and remove the matching rows. We also count variety markers in the outputs: \emph{vosotros} forms,
\emph{ustedes}, voseo, and curated lists of Spain-only and Latin America-only words for Spanish; marker words for
Asturian, Aragonese, Aranese and Languedocien.

\section{Castilian and Latin American Spanish}
\label{sec:spanish}

\subsection{Finding variety text in a web corpus}

Web corpora rarely say where a text was written. We indexed all 5,000 shards of MADLAD-400's Spanish
\texttt{clean} split: 250.9M documents, 1.06\,TB of text, 5.0M websites. As weak labels we use the website's
domain: \texttt{.es} for Spain and the Latin American country domains for Latin America (\texttt{.co} excluded, as
it is widely used as a generic domain).

\paragraph{Variety classifier.} A supervised fastText classifier \citep{joulin2017bag} (64 dimensions, word bigrams)
trained on 912k documents per class from 200 shards, at most 200 documents per website, on text normalised to
lowercase content words (capitalised words, digits and URLs removed, which hides most place names and brands). On
5\% held-out websites, against the domain labels, it reaches precision 0.963 and recall 0.743 for Spain at a
threshold of 0.9 (0.917 / 0.928 at 0.5). An independent check with grammatical markers confirms that it follows how
the text is written: documents scored $\geq 0.9$ contain \emph{vosotros} markers in 175 and voseo in 5 per 1,000,
against 12 and 32 per 1,000 for documents scored $< 0.1$. The classifier is reliable on documents and websites
(averaged over websites, only 0.2--0.3\% land on the opposite side), but close to random on single sentences.

\paragraph{Selection rule.} Members of our team labelled an internal reference set of 340 documents, stratified by
domain group and classifier score, as Latin American or not. Table~\ref{tab:selection} compares candidate rules. The
adopted rule E keeps documents from Spanish domains or from generic-domain websites with at least 20 documents whose
mean Spain score is $\geq 0.85$ (241,801 websites, e.g.\ national newspapers on \texttt{.com}), with a document score
$\geq 0.9$ and no voseo or Latin American vocabulary markers. None of its 72 reference documents is Latin American,
which bounds the Latin American share at roughly 4\% (95\% confidence). The Latin American rule mirrors it: Latin
American country domains or generic websites with a mean score $\leq 0.15$, document score $\leq 0.1$, and no
\emph{vosotros}, Spain vocabulary or voseo (for a neutral variety); it selects 61.9M documents.

\begin{table}[t]
\centering\small
\resizebox{\textwidth}{!}{%
\begin{tabular}{lrrrr}
\toprule
Rule & Ref.\ docs & Latin American & Est.\ docs & Est.\ Latin American \\
\midrule
A. Spain domains, all & 116 & 19 & 39.8M & 5.2\% \\
B. Spain domains, score $\geq 0.9$ & 48 & 0 & 29.3M & 0\% \\
F. Score $\geq 0.9$, any domain & 136 & 16 & 92.9M & 4.5\% \\
\textbf{E. Spain domains or allowlisted sites, score $\geq 0.9$, no markers} & 72 & 0 & 71.1M & 0\% \\
\bottomrule
\end{tabular}}
\caption{Candidate rules for selecting Castilian documents, on the 340-document reference set (estimates weighted to
the whole split).}
\label{tab:selection}
\end{table}

\paragraph{Sentences.} From 100 random shards we extract prose sentences of 6--60 words, removing boilerplate, spam,
non-Spanish lines and sentences with markers of the other variety, then deduplicate and cap at 40 sentences per
document and 5,000 per website. This gives 18.7M Castilian sentences (438M words) and 13.0M Latin American sentences
(349M words), with development splits that hold out whole websites. For Latin America we also cap each country domain
at 15\% of the selection, so Mexico, Argentina and Chile do not dominate.

\paragraph{Back-translated pairs.} We back-translate 1.2M sentences per variety into eight languages (English 70\%,
French, Portuguese, Catalan, Italian, German, Galician, Basque), keeping 98.4--98.6\%: 1,180,728 Castilian and
1,183,613 Latin American pairs. All Castilian sentences with variety features were favoured (30\% of the Castilian
selection), but only 5.6\% of the Latin American selection has an explicit feature, as web news rarely mentions
peanuts or parking.

\paragraph{Human pairs.} Versions 3 and 4 of the Castilian model add human English--Spanish pairs from OPUS
\citep{tiedemann2012opus}: Europarl v8 \citep{koehn2005europarl} (120,000 pairs, Spain's EU Spanish) and Tatoeba
\citep{tatoeba} (185--193k), filtered for Latin American markers and \emph{ustedes}. The Latin American model uses
the Tatoeba pairs without \emph{vosotros}, voseo or Spain vocabulary (192,292 pairs), and no Europarl.

\subsection{Versions}

We trained four Castilian versions (Table~\ref{tab:es-versions}) and one Latin American version. v1 uses only the
back-translated pairs. v2 halves the learning rate and raises replay to 40\% to reduce forgetting; it does not
(Section~\ref{sec:forgetting}) and was not released. v3 replaces part of the back-translated data with human pairs,
which fixes literal renderings of English idioms but loses \emph{vosotros} for informal plural address, because
Tatoeba's translations often use \emph{ustedes}. v4 combines all back-translated pairs with the human pairs and three
filters: Tatoeba pairs whose ``you'' sentences show no \emph{tú}, \emph{vosotros} or \emph{usted} cue are dropped;
back-translated pairs whose English has no ``you'' but whose Spanish addresses the reader with \emph{vosotros} or
\emph{os} (blog style, ``os comentamos'') are dropped (18,964), as they teach the model to add it; and pairs with
garbled characters are dropped. The Latin American model follows v4's recipe with 1,375,905 pairs.

\begin{table}[t]
\centering\small
\begin{tabular}{llrrr}
\toprule
Version & Data (+ 25\% replay unless noted) & Steps & Hours & Val.\ loss \\
\midrule
es\_ES v1 & 1.18M back-translated & 12,258 & 17.5 & 0.322 \\
es\_ES v2 & 0.80M back-translated, 40\% replay, lr $10^{-4}$ & 10,411 & 11.2 & 0.327 \\
es\_ES v3 & 0.84M back-translated + 0.31M human & 11,975 & 14.3 & 0.370$^\dagger$ \\
es\_ES v4 & 1.16M back-translated + 0.31M human, filtered & 15,256 & 22.0 & 0.366$^\dagger$ \\
es\_419 & 1.18M back-translated + 0.19M human (Tatoeba) & 14,328 & 17.5 & 0.321$^\ddagger$ \\
\bottomrule
\end{tabular}
\caption{Spanish variety runs. Validation sets differ: $^\dagger$ adds the human dev sets; $^\ddagger$ Latin
American dev set.}
\label{tab:es-versions}
\end{table}

\subsection{Results}

\paragraph{Targeted tests.} We wrote short English test sentences for each variety decision: informal plural
\emph{you} (\emph{you guys, kids}\dots), formal address, everyday vocabulary where the varieties differ
(\emph{phone, computer, juice, parking lot}\dots), borderline vocabulary used in both (\emph{take the bus, sweater},
or \emph{car, glasses} for Latin America), and control sentences without \emph{you} (no \emph{vosotros} should
appear). Table~\ref{tab:targeted} gives the share of outputs with the expected choice (97 sentences for Castilian,
89 for Latin America).

\begin{table}[t]
\centering\small
\begin{tabular}{lrrrrr}
\toprule
& Informal plural & Formal & Vocabulary & Borderline & Controls \\
\midrule
\multicolumn{6}{l}{\emph{Castilian choices}} \\
Base \ltag{es} & 10\% & 100\% & 62\% & 0\% & 100\% \\
es\_ES v1 & \textbf{78\%} & 73\% & \textbf{97\%} & \textbf{86\%} & 70\% \\
es\_ES v3 & 32\% & 73\% & 90\% & 0\% & 70\% \\
es\_ES v4 & 70\% & 73\% & \textbf{97\%} & 14\% & 70\% \\
\midrule
\multicolumn{6}{l}{\emph{Latin American choices}} \\
Base \ltag{es} & 72\% & 100\% & 74\% & 40\% & 100\% \\
es\_419 & \textbf{80\%} & 100\% & \textbf{83\%} & 20\% & 100\% \\
\bottomrule
\end{tabular}
\caption{Targeted tests: share of outputs with the expected variety choice. The Latin American informal-plural score
is a lower bound: 5 of its 8 misses are valid \emph{ustedes} forms the test does not list.}
\label{tab:targeted}
\end{table}

\paragraph{Human references.} On 1,000 held-out human-translated sentences each (Table~\ref{tab:human}), v4 and v3
are slightly better than the base model on Europarl and clearly better on Tatoeba, while v1, trained only on
back-translated web text, is below the base model. Part of the gain of v3, v4 and es\_419 is domain match, as they
were trained on other pairs of these corpora.

\begin{table}[t]
\centering\small
\begin{tabular}{lccc}
\toprule
& Europarl (es\_ES refs.) & Tatoeba (es\_ES refs.) & Tatoeba (es\_419 refs.) \\
\midrule
Base \ltag{es} & 65.5 / 44.2 & 72.8 / 56.3 & 73.0 / 57.1 \\
es\_ES v1 & 64.6 / 43.1 & 69.9 / 52.0 & -- \\
es\_ES v3 & 65.8 / 44.8 & 74.6 / 59.3 & -- \\
es\_ES v4 & 65.8 / 44.8 & 74.6 / 59.1 & -- \\
es\_419 & -- & -- & 74.6 / 59.5 \\
\bottomrule
\end{tabular}
\caption{Human-reference test sets (English$\to$Spanish), chrF++ / BLEU.}
\label{tab:human}
\end{table}

\paragraph{FLORES.} On FLORES-200 English$\to$Spanish, chrF++ goes from 54.1 to 53.4 (es\_ES v4) and 53.5 (es\_419).
FLORES sentences are formal and rarely show the variety, and its Spanish reference leans slightly Latin American
(0.8\% of sentences with Latin American vocabulary, 0.2\% with Spain vocabulary). The Castilian model removes Latin
American vocabulary (1.0\% $\to$ 0.1\%) and the Latin American model removes Spain vocabulary (0.2\% $\to$ 0.0\%);
part of the chrF++ difference comes from words that now differ from the reference.

\paragraph{In-domain sets.} On about 2,100 English sentences back-translated from held-out sentences of each variety
(favouring the fine-tunes), chrF against the original sentence rises from 78.0 to 80.6 (Castilian) and from 78.2 to
80.3 (Latin American), and the variety profile moves to the original's: \emph{vosotros} in 17.6\% of Castilian
outputs (base 7.2\%, original 21.2\%), Latin American vocabulary in 12.4\% of Latin American outputs (base 6.9\%,
original 15.8\%).

\subsection{Tag collapse}
\label{sec:collapse}

In every Spanish run, plain \ltag{es} drifted to the new variety. On the in-domain sets the parent and the new tag
give identical translations for 87.0\% (es\_ES v4) and 84.1\% (es\_419) of sentences, with the same variety
profile. The replay pairs with \ltag{es} targets are mostly neutral Spanish and do not teach the model to tell the
tags apart. We accepted this for the Spanish models and state it on their model cards: users who want the base
model's Spanish should use the base model. For the Iberian model, where \ltag{oc} and \ltag{es} must keep producing
standard Occitan and Spanish, we added the contrastive pairs and the separate tag learning rate of
Section~\ref{sec:method}; Section~\ref{sec:separation} shows that the tags stay apart.

\section{Asturian, Aragonese and Aranese}
\label{sec:iberian}

\subsection{Data}

All the training data of this model is publicly available. Table~\ref{tab:ib-data} lists the training pairs.

\paragraph{Asturian and Aragonese.} The TAN-IBE corpora \citep{tanibe_ast,tanibe_arg} contain Spanish--Asturian and
Spanish--Aragonese sentence pairs found in Wikipedia, monolingual Wikipedia text, Apertium back-translations of the
monolingual text, and Apertium translations of the Spanish side of the NTEU English--Spanish corpus. FastText
\texttt{lid.176} labels Asturian and Aragonese as Spanish, so we identify them with marker words: at least two
Asturian markers (\emph{ye, nun, tamién, dende, pela, colos, güei}\dots), or at least two Aragonese markers
(\emph{d'o, d'a, ta, dica, ixo, chent, luenga}\dots) with at most one Spanish-only word and no Aranese marker. On
FLORES+ dev these rules keep 55\% of Asturian sentences and no Spanish ones. The human Asturian and Aragonese
sentences are back-translated by the base model: Asturian into one of Spanish, Catalan, English and French; Aragonese,
whose clean Wikipedia text is small (45k sentences), into Spanish and two of the other three. Wikipedia's Spanish
sentences give a human source for the aligned pairs. In the second run (ib2) we also recover 1,383 aligned
Aragonese pairs that the first filters dropped (one marker suffices for text that TAN-IBE already aligned as
Aragonese), count the 3,408 human Spanish--Aragonese pairs four times, and add 11,498 Aragonese Wikipedia sentences
with TAN-IBE's Apertium Spanish as source. Apertium Asturian and Aragonese, being machine output, are used only as
sources, for translation into Spanish (40,000 sentences per language).

\paragraph{Aranese.} The BSC Catalan--Aranese corpus \citep{bsc_aranese}, PILAR \citep{pilar}, the Projecte Aina
Spanish--Aranese corpus \citep{aina_esoc}, Aranese documents of MADLAD-400, Occitan Wikipedia and Common Voice texts,
and TAN-IBE's Aranese Wikipedia sentences with their Spanish \citep{tanibe_arn}, back-translated into Spanish,
Catalan, English and French; human standard Occitan with \ltag{oc} (30,000 pairs); and contrastive pairs with the
base model's \ltag{oc} output.

\begin{table}[t]
\centering\small
\resizebox{\textwidth}{!}{%
\begin{tabular}{lrl}
\toprule
Part & Pairs & Content \\
\midrule
Aranese, X$\to$arn & 450,246 & back-translated human Aranese; Spanish--Aranese pairs \\
Aranese$\to$X & 100,000 & incl.\ 40k Apertium Aranese $\to$ human Spanish \\
Standard Occitan & 30,000 & human Occitan targets, \ltag{oc} \\
Aranese contrastive & 160,000 & 80k sources $\times$ (\ltag{oc\_aranes}, \ltag{oc}) \\
Asturian, X$\to$ast & 446,178 & Wikipedia es--ast pairs; back-translated human Asturian \\
Asturian contrastive & 80,000 & \ltag{ast} human vs.\ \ltag{es} Spanish \\
Asturian$\to$Spanish & 80,000 & Wikipedia pairs; 40k Apertium Asturian sources \\
Aragonese, X$\to$an & 164,130 & 3,408 human pairs $\times$4; back-translated; Apertium Spanish sources \\
Aragonese contrastive & 60,000 & \ltag{an} human vs.\ \ltag{es} Spanish \\
Aragonese$\to$Spanish & 43,408 & Wikipedia pairs; 40k Apertium Aragonese sources \\
\midrule
Total & 1,613,962 & + 537,987 replay pairs \\
\bottomrule
\end{tabular}}
\caption{Training pairs of the Iberian model (ib2).}
\label{tab:ib-data}
\end{table}

\paragraph{Training.} Both runs start from the base model with two new tags (\ltag{oc\_aranes} from \ltag{oc},
\ltag{ast} from \ltag{es}), tag learning rate $2\cdot10^{-3}$, replay 25\%. ib1 trained 16,543 steps and ib2 16,798
steps (19.3 hours); validation loss 0.565 and 0.560 on the same development set.

\subsection{Results}

We ran every system ourselves on FLORES+ devtest (1,012 sentences), with the decoding of its model card: TAN-IBE's
Marian models converted to CTranslate2 (beam 12), SalamandraTA-7B and 2B \citep{garcia-gilabert-etal-2025-salamandra} with their chat prompt (beam 5), IbRo-nllb
(beam 5) and the base model. Table~\ref{tab:ib-flores} gives the Spanish directions and Table~\ref{tab:ib-other} the
directions from and into Catalan, English and French.

\begin{table}[t]
\centering\small
\resizebox{\textwidth}{!}{%
\begin{tabular}{lcccccc}
\toprule
& es$\to$ast & ast$\to$es & es$\to$arg & arg$\to$es & es$\to$arn & arn$\to$es \\
\midrule
TAN-IBE (Marian) & 17.8 / 51.3 & 20.7 / 49.2 & \textbf{60.4 / 78.8} & 63.8 / 78.5 & 50.9 / 73.0 & 51.6 / 71.8 \\
SalamandraTA-7B & \textbf{24.3 / 54.7} & 23.8 / 51.9 & 59.1 / 78.2 & \textbf{69.6 / 81.5} & \textbf{54.2 / 75.0} & \textbf{63.0 / 77.7} \\
SalamandraTA-2B & 17.9 / 51.6 & \textbf{24.7 / 52.3} & 39.0 / 68.0 & 60.8 / 76.3 & 45.6 / 70.3 & 54.4 / 72.3 \\
IbRo-nllb & 16.8 / 50.9 & 21.3 / 49.9 & 59.9 / 78.8 & 65.3 / 78.8 & 50.3 / 73.2 & 56.4 / 73.3 \\
MADLAD-400-3B & -- & 21.2 / 49.5 & 35.0 / 67.7 & 65.5 / 79.1 & 10.8 / 46.1$^*$ & 55.4 / 72.6 \\
\midrule
ib1 & 17.8 / 51.5 & 21.5 / 50.1 & 43.4 / 72.2 & 68.0 / 80.4 & 48.8 / 72.3 & 61.6 / 76.7 \\
ib2 & 17.8 / 51.5 & 21.5 / 50.1 & 44.0 / 72.7 & 68.1 / 80.4 & 49.0 / 72.5 & 61.2 / 76.5 \\
\bottomrule
\end{tabular}}
\caption{FLORES+ devtest, Spanish directions, BLEU / chrF. $^*$\ltag{oc}, standard Occitan.}
\label{tab:ib-flores}
\end{table}

\begin{table}[t]
\centering\small
\resizebox{\textwidth}{!}{%
\begin{tabular}{lcccccc}
\toprule
& ca/en/fr$\to$ast & ast$\to$ca/en/fr & ca/en/fr$\to$arg & arg$\to$ca/en/fr & ca/en/fr$\to$arn & arn$\to$ca/en/fr \\
\midrule
SalamandraTA-7B & \textbf{31.0/35.7/29.0} & \textbf{36.3/45.2/39.1} & \textbf{27.2/27.0/23.6} & \textbf{33.4/36.6/33.7} & \textbf{33.0/27.5/23.9} & \textbf{41.1/36.1/33.5} \\
SalamandraTA-2B & 25.5/27.4/22.4 & 34.8/40.9/35.1 & 17.8/16.3/14.7 & 31.6/32.6/30.3 & 29.5/20.3/17.5 & 38.9/31.8/30.8 \\
MADLAD-400-3B & -- & 31.0/37.3/32.9 & 16.4/15.0/13.4 & 29.5/31.1/30.0 & 8.5/6.8/5.8$^*$ & 36.9/28.3/28.4 \\
ib2 & 27.7/29.9/24.5 & 32.5/39.0/34.6 & 19.6/17.4/15.8 & 30.4/30.4/30.1 & 31.3/22.2/20.3 & 38.7/31.0/31.4 \\
\bottomrule
\end{tabular}}
\caption{FLORES+ devtest, Catalan, English and French directions, BLEU. $^*$\ltag{oc}.}
\label{tab:ib-other}
\end{table}

Against TAN-IBE, ib2 ties Spanish$\to$Asturian, is ahead in Asturian$\to$Spanish (+0.8 BLEU), Aragonese$\to$Spanish
(+4.3) and Aranese$\to$Spanish (+9.6), and slightly behind in Spanish$\to$Aranese ($-1.9$). SalamandraTA-7B, more
than twice our size, is the best system almost everywhere. Into Asturian, Aragonese and Aranese from Catalan,
English and French, ib2 is second after it, ahead of SalamandraTA-2B and the base model. Spanish$\leftrightarrow$Asturian
scores are low for every system: the FLORES+ Asturian sentences are free translations. On a held-out test of 1,000
BSC Catalan--Aranese pairs, ib2 reaches 51.9 / 73.3 into Aranese (SalamandraTA-7B 46.7 / 70.3), but this test is in
our training domain.

\subsection{Each tag keeps its language}
\label{sec:separation}

Table~\ref{tab:separation} translates the 1,012 English FLORES+ sentences with each tag. No output of a new tag is
identical to its parent's. \ltag{oc\_aranes} outputs contain Aranese words in 79\% of sentences (reference 80\%) and
Languedocien words in 1.3\%, while \ltag{oc} keeps standard Occitan (Languedocien words in 91\%) and gains 1.1 chrF.
\ltag{ast} meets the Asturian marker rule in 61\% of outputs (reference 58\%), \ltag{es} in 1\%. Spanish loses 0.5
chrF, in line with the forgetting of Section~\ref{sec:forgetting}. Into Spanish from Asturian or Aragonese, 1\% of the
outputs look Asturian or Aragonese, as often as the Spanish reference does. One failure remains: \ltag{es} applied
to Spanish input, a case absent from training, can answer in Asturian.

\begin{table}[t]
\centering\small
\begin{tabular}{llccc}
\toprule
Tag & Model & chrF & Variety markers & Identical to parent \\
\midrule
\ltag{oc} & base / ib2 & 63.7 / 64.8 & Languedocien 87\% / 91\% & -- \\
\ltag{oc\_aranes} & ib2 & 53.4 & Aranese 79\%, Languedocien 1.3\% & 0\% \\
\ltag{es} & base / ib2 & 56.4 / 55.9 & Asturian rule 0.8\% / 1.1\% & -- \\
\ltag{ast} & ib2 & 60.8 & Asturian rule 61\% & 0.1\% \\
\ltag{an} & base / ib2 & 48.8 / 51.6 & Aragonese rule 67\% / 74\% & 0\% \\
\bottomrule
\end{tabular}
\caption{English$\to$X on FLORES+ devtest with each tag. chrF against the reference of the tag's language; share of
outputs with variety markers; share identical to the parent tag's output (\ltag{oc} for \ltag{oc\_aranes}, \ltag{es}
for \ltag{ast} and \ltag{an}).}
\label{tab:separation}
\end{table}

\subsection{Spanish$\to$Aragonese: a difference of written variety}
\label{sec:variety-mismatch}

ib2 is 16 BLEU behind TAN-IBE and IbRo-nllb in Spanish$\to$Aragonese, while it is close to or ahead of them in all
other Spanish directions. Adding human Aragonese pairs between ib1 and ib2 changed little (+0.6 BLEU). Counting words
in the 1,012 outputs shows why (Table~\ref{tab:arg-words}): the FLORES+ Aragonese reference writes the articles
\emph{lo, la, los, las} and \emph{per, pa, puede}, whereas Aragonese Wikipedia, the source of all our human
Aragonese, mostly writes \emph{o, a, os, as} and \emph{por, ta, puet}, and so does our model. TAN-IBE and IbRo-nllb,
trained largely on Apertium output, match the reference almost word for word. On 224 held-out Spanish--Aragonese
Wikipedia pairs the ranking reverses: ib2 50.9 / 73.3, ib1 50.4 / 72.8, TAN-IBE 45.0 / 70.2, IbRo-nllb 42.6 / 69.4.
That test is in our training domain and may be in TAN-IBE's training data, so neither test settles which system is
better; each favours one written variety. Both varieties are written Aragonese. Benchmarks for languages with
several written standards or strong regional variation should document which one their references follow, as a
single reference turns a choice of variety into a difference in score.

\begin{table}[t]
\centering\small
\begin{tabular}{lcccc}
\toprule
& \emph{lo / la / los / las} & \emph{o / os / as} & \emph{per / pa} & \emph{por / ta} \\
\midrule
FLORES+ reference & 693 / 650 / 446 / 301 & 150 / 43 / 47 & 242 / 199 & 0 / 59 \\
TAN-IBE & 785 / 639 / 444 / 291 & 156 / 53 / 51 & 233 / 201 & 2 / 4 \\
IbRo-nllb & 794 / 637 / 446 / 293 & 157 / 49 / 49 & 237 / 203 & 0 / 0 \\
ib2 & 143 / 108 / 38 / 53 & 762 / 467 / 284 & 0 / 3 & 231 / 209 \\
\bottomrule
\end{tabular}
\caption{Word counts in the 1,012 Spanish$\to$Aragonese FLORES+ devtest outputs.}
\label{tab:arg-words}
\end{table}

\section{Forgetting}
\label{sec:forgetting}

Every fine-tune loses a little in the thirteen other FLORES-200 directions (Table~\ref{tab:forgetting}), most in
English$\to$Japanese and English$\to$Chinese (up to $-2.0$ chrF++), least in German$\to$English and
English$\to$Hindi. The loss is remarkably constant across very different training data, from two Spanish varieties
to three minority languages, and halving the learning rate while raising replay to 40\% (es\_ES v2) reduced it by
only 0.05. Replay with the base model's beam-search outputs keeps the behaviour but not the exact quality of the
base model; distillation against the base model's distributions, or decoder-only adaptation, are options we have not
tried.

\begin{table}[t]
\centering\small
\begin{tabular}{lccccccc}
\toprule
& es\_ES v1 & es\_ES v2 & es\_ES v3 & es\_ES v4 & es\_419 & ib1 & ib2 \\
\midrule
Mean $\Delta$ chrF++ & $-0.64$ & $-0.59$ & $-0.58$ & $-0.62$ & $-0.65$ & $-0.62$ & $-0.69$ \\
\bottomrule
\end{tabular}
\caption{Mean chrF++ change over thirteen FLORES-200 devtest directions against the base model.}
\label{tab:forgetting}
\end{table}

\section{Limitations}

Evaluation is automatic, with single references, and the outputs were not reviewed systematically by native
speakers; the targeted tests are small (89--97 sentences). The Spanish ``varieties'' are coarse: Castilian means
written in Spain, including Canarian and Andalusian usage, and the Latin American model aims at a neutral Spanish
whose vocabulary mixes countries. The variety selection is automatic and document-level, and its precision is
bounded on only 72 reference documents. Most training sources are machine translations. In the Iberian model,
Asturian and Aragonese come almost entirely from Wikipedia, and the Aragonese data follows the norm of the Academia
Aragonesa de la Lengua; outputs may not follow the norms of the Academia de la Llingua Asturiana or the Institut
d'Estudis Aranesi. We observed Spanish outputs for some short informal sentences with \ltag{ast}, and meaning errors
such as a wrong grammatical person.

\section{Release}

All models are released under Apache-2.0, the licence of the base model, with CTranslate2 int8 versions; the
datasets under ODC-BY 1.0, MADLAD-400's licence:
\begin{itemize}
\item \texttt{Ugiat/madlad400-3b-mt-es\_ES} (v4 on \texttt{main}, v1 and v3 on branches);
\item \texttt{Ugiat/madlad400-3b-mt-es\_419};
\item \texttt{Ugiat/madlad400-3b-mt-ast-an-oc\_aranes} (ib2);
\item \texttt{Ugiat/spanish-variety-fasttext}, the variety classifier;
\item \texttt{Ugiat/madlad-400-es-annotations}, a document-level index of MADLAD-400's Spanish split (domain group,
  variety score, languages of the lines, selection flags);
\item \texttt{Ugiat/madlad-400-es\_ES-sentences}, \texttt{Ugiat/madlad-400-es\_419-sentences} and the corresponding
  back-translated pairs.
\end{itemize}

\section*{Use of AI assistants}

During a pilot, Claude Haiku and Claude Sonnet labelled the 340 reference documents of Section~\ref{sec:spanish} as
Spain or Latin America, to compare with the classifier and the selection rules; these labels were not used to select
or create training data. No training data of any model in this report was generated or edited by a large language
model. The code, the data pipelines and a first draft of this report were written with the help of an AI coding
assistant (Claude Code), and were checked by the authors.

\section*{Acknowledgements}

We thank the TAN-IBE project and Antoni Oliver for publishing their corpora and models, and the authors of
MADLAD-400, PILAR, the BSC Catalan--Aranese corpus, the Projecte Aina corpora, SalamandraTA, IbRo-nllb, Tatoeba,
Europarl and FLORES+.

\bibliographystyle{unsrtnat}
\bibliography{references}

\end{document}